\documentclass[sigconf]{acmart}
\AtBeginDocument{%
  }
\usepackage{multirow}
\usepackage{makecell}
\usepackage{ulem}
\usepackage{color} 
\def\etal{et al.}

\setcopyright{acmlicensed}

\copyrightyear{2025}
\acmYear{2025}
\setcopyright{acmlicensed}\acmConference[MM '25]{Proceedings of the 33rd ACM International Conference on Multimedia}{October 27--31, 2025}{Dublin, Ireland}
\acmBooktitle{Proceedings of the 33rd ACM International Conference on Multimedia (MM '25), October 27--31, 2025, Dublin, Ireland}
\acmDOI{10.1145/3746027.3755291}
\acmISBN{979-8-4007-2035-2/2025/10}

\begin{document}

\title[SAM-TTT: Segment Anything Model via Reverse \\Parameter Configuration and Test-Time Training]{SAM-TTT: Segment Anything Model via Reverse Parameter Configuration and Test-Time Training for Camouflaged Object Detection}


\author{Zhenni Yu}
\affiliation{%
  \institution{College of Computer Science and Artificial Intelligence, Wenzhou University}
  \city{Wenzhou}
  \state{Zhejiang}
  \country{China}
}
\affiliation{%
  \institution{College of Computer Science and Technology, Tongji University}
  \city{Shanghai}
  \country{China}
}
\email{zhenn_yu@163.com}

\author{Li Zhao}
\affiliation{%
  \institution{College of Information Science and Technology, Zhejiang Shuren University}
  \city{Hangzhou}
  \state{Zhejiang}
  \country{China}}
\email{zhaoli@zjsru.edu.cn}

\author{Guobao Xiao}
\authornote{corresponding authors.}
\affiliation{%
  \institution{College of Computer Science and Technology, Tongji University}
  \city{Shanghai}
  \country{China}
}
\email{x-gb@163.com}

\author{Xiaoqin Zhang}
\authornotemark[1]
\affiliation{%
 \institution{College of Computer Science and Artificial Intelligence, Wenzhou University}
  \city{Wenzhou}
  \state{Zhejiang}
  \country{China}
}
\email{zhangxiaoqinnan@gmail.com}

\renewcommand{\shortauthors}{Zhenni Yu et al.}

\begin{abstract}
This paper introduces a new Segment Anything Model (SAM) that leverages reverse parameter configuration and test-time training to enhance its performance on Camouflaged Object Detection (COD), named SAM-TTT. While most existing SAM-based COD models primarily focus on enhancing SAM by extracting favorable features and amplifying its advantageous parameters, a crucial gap is identified: insufficient attention to adverse parameters that impair SAM's semantic understanding in downstream tasks. To tackle this issue, the Reverse SAM Parameter Configuration Module is proposed to effectively mitigate the influence of adverse parameters in a train-free manner by configuring SAM's parameters. Building on this foundation, the T-Visioner Module is unveiled to strengthen advantageous parameters by integrating Test-Time Training layers, originally developed for language tasks, into vision tasks. Test-Time Training layers represent a new class of sequence modeling layers characterized by linear complexity and an expressive hidden state. By integrating two modules, SAM-TTT simultaneously suppresses adverse parameters while reinforcing advantageous ones, significantly improving SAM's semantic understanding in COD task. Our experimental results on various COD benchmarks demonstrate that the proposed approach achieves state-of-the-art performance, setting a new benchmark in the field. The code will be available at https://github.com/guobaoxiao/SAM-TTT.
\end{abstract}

\ccsdesc[500]{Computing methodologies~Object detection}
\ccsdesc[300]{Computing methodologies~Computer vision}
\ccsdesc[100]{Computing methodologies~Artificial intelligence}

\keywords{Camouflaged Object Detection, Segment Anything Model, Test-Time Training, Parameter Configuration}


\maketitle

\section{Introduction}
\label{sec:intro}
\begin{figure}[t!]
\centering{\includegraphics[width=0.8\linewidth]{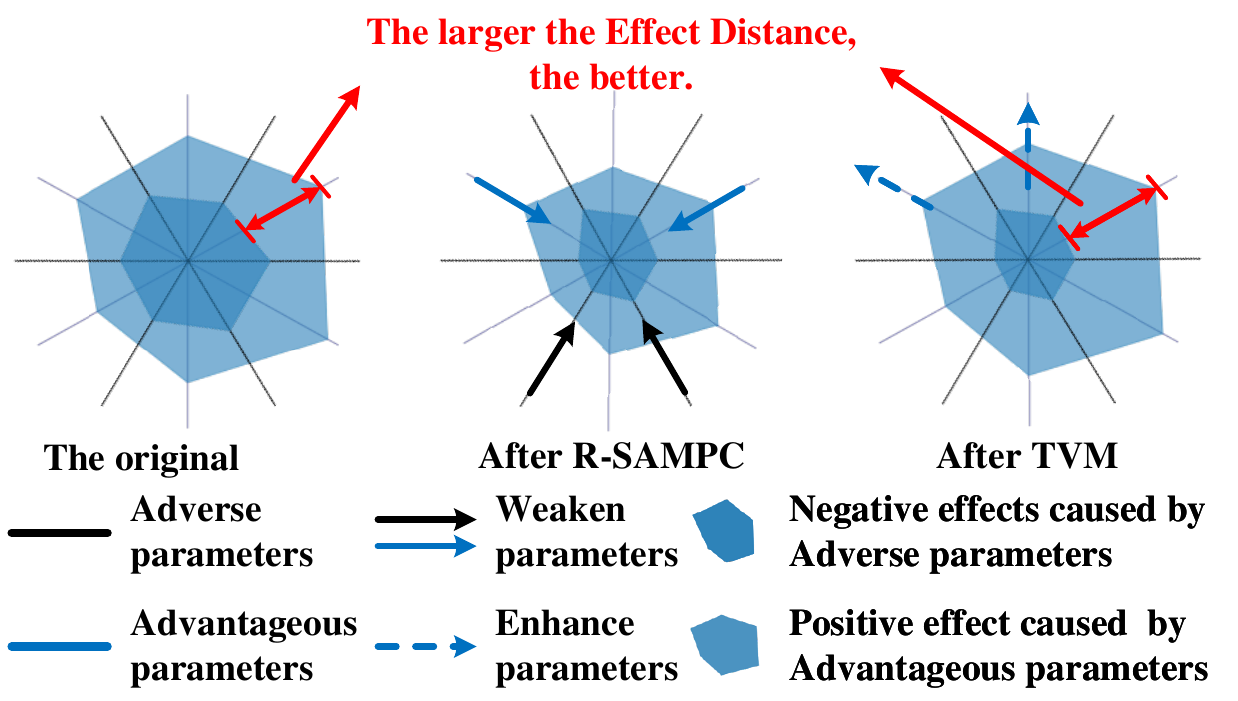}}
\caption{\small{\textbf{Conceptual Diagram of SAM-TTT from the Perspective of Effect Distance.} The roles of the R-SAMPC and TVM. R-SAMPC weakens the adverse and advantageous parameters, while TVM strengthens the advantageous parameters.}}
\label{fig:begin}
\end{figure}
\begin{figure}[h]
\centering{\includegraphics[width=0.8\linewidth]{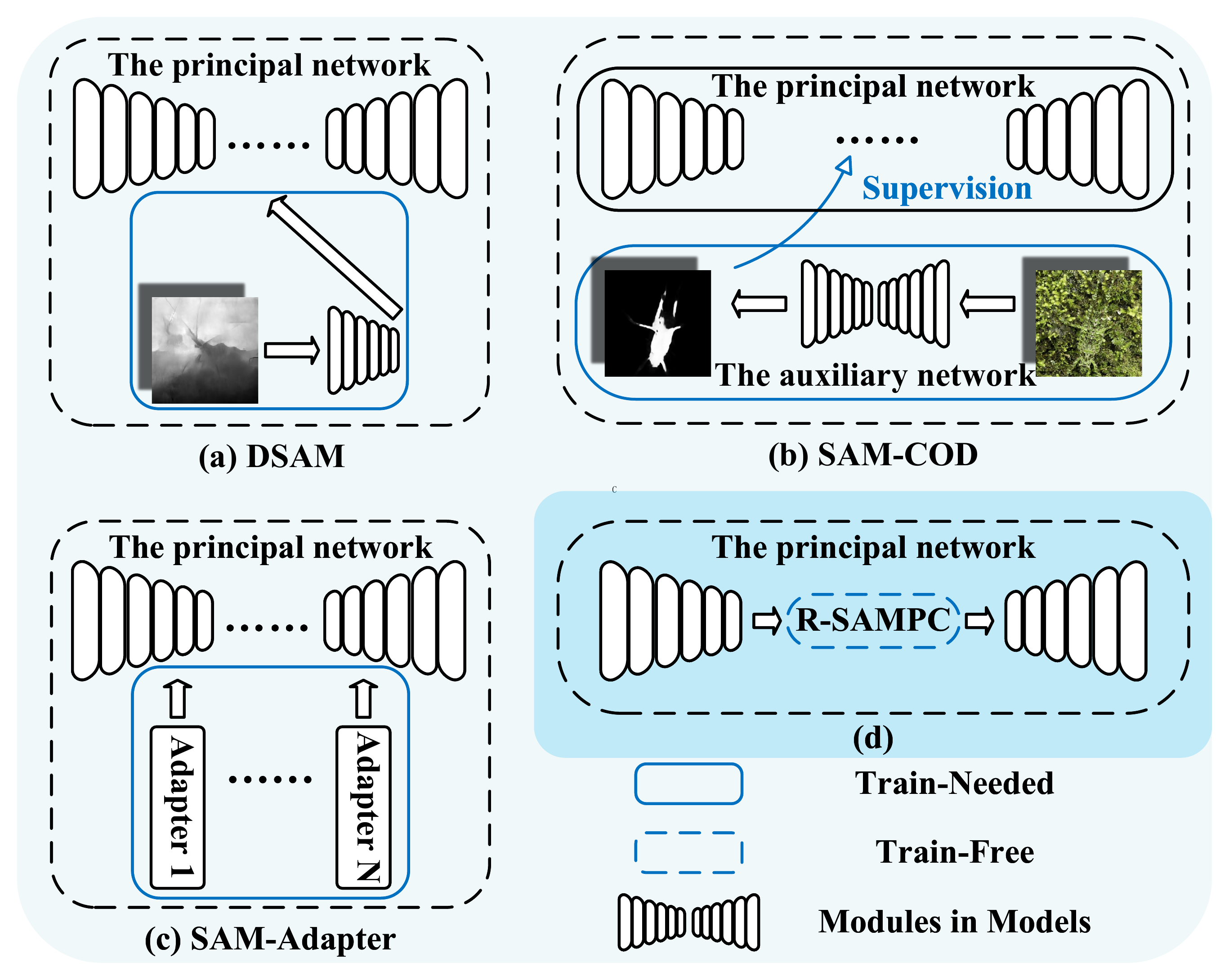}}
\vspace{-5pt}
\caption{\small{\textbf{A comparative analysis of solutions to mitigate the semantic deficiency encountered when applying SAM to COD.}}}
\label{fig:dropout}
\end{figure}
Segment Anything Model (SAM)~\cite{kirillov2023segment} is undergoing rapid development. Its impressive zero-shot capabilities enable good results in various downstream tasks while also saving training costs, leading to widespread utilization and exploration. Currently, SAM faces the issue of semantic deficiency as it progresses towards downstream tasks~\cite{ma2024segment}. In the specific downstream task of Camouflaged Object Detection (COD), semantic deficiency issues manifest in the segmentation masks not aligning with the intended semantics~\cite{ji2023sam}. This phenomenon suggests that SAM produces object-biased fine-grained semantic responses in COD~\cite{chen2024sam}. The underlying cause is SAM's zero-shot capabilities stem from the SA-1B dataset. The SA-1B dataset exhibits a domain gap compared to the COD dataset. 
Several studies have been conducted to address this issue. SAM-COD~\cite{chen2024sam} produces a preliminary mask using additional network, which contains an initial understanding, to correct the above errors by calculating the semantic entropy. SAM-Adapter~\cite{chen2023sam} utilizes adapter technology to integrates domain knowledge to SAM. Aware of the limitations of RGB modality, DSAM~\cite{yu2024exploring} adds depth information to correct the semantic missing. 
However, all these methods require the integration of additional coding modules to fulfill their objectives, which adds complexity to their implementation. A deeper underlying reason is that most researchers focus on introducing semantic information to compensate for its absence, while few pay attention to the adverse parameters within SAM that contribute to the generation of erroneous semantic information.

To solve these problems, a new SAM-based model is proposed for COD, named SAM-TTT. SAM-TTT integrates the Reverse SAM Parameter Configuration Module (R-SAMPC) and T-vision Module (TVM) to strengthen SAM's performance on COD. SAM-TTT focuses on the impact of parameters in SAM and introduces a conceptual metric called \textbf{Effect Distance}, which represents the quality of parameters—it increases as beneficial parameters are enhanced and detrimental ones are suppressed. SAM-TTT configures the parameters of SAM in a new way of widening the Effect Distance, shown in Figure~\ref{fig:begin}. This concept derives from the unique working mechanism of R-SAMPC. Unlike existing approaches that use additional coding modules to introduce semantic information, as illustrated in (a), (b), (c) of Figure~\ref{fig:dropout}, R-SAMPC solves the problem in a train-free way, as shown in (d) of Figure~\ref{fig:dropout}. It is through a designed randomly initialized convolution module that does not update to loosen the constraints on the parameters, ultimately reducing the impact of adverse parameters in SAM for COD. R-SAMPC configures the parameters in this novel way and mitigates the semantic deficiency caused by adverse parameters. It's used during training but not during inference. R-SAMPC focuses on weakening the parameters, which is contrary to the previous approaches that emphasized enhancing the parameters; thus, it is named ``Reverse". From another perspective, R-SAMPC is viewed as a random mask at the parameter level, corresponding to random initialization. The random mask is able to suppress strong responses~\cite{kumar2017hide}. Chen~\etal~\cite{chen2024just} diverted the model's attention across the entire object by partially occluding the labeled regions, rather than focusing on the most distinctive parts. R-SAMPC disturbs the part of the model that is highly responsive in a train-free way and forces the model to focus on the whole. R-SAMPC addresses the semantic deficiency issues caused by SAM in COD by weakening the adverse parameters in a train-free manner. 

Furthermore, R-SAMPC introduces a certain level of disruption for advantageous parameters. TVM is proposed to extract favorable features to compensate for this interference. TVM adopts TTT-Linear, an instantiation of the TTT layer~\cite{sun2024learning}. TTT is an RNN layer with linear complexity and a highly expressive hidden state. TVM incorporates TTT into computer vision to extract advantageous features. TTT lets the hidden state itself be a weight. This approach breaks the restriction that their representation in long context is limited by the expressive power of hidden states~\cite{sun2024learning}. 

SAM-TTT proposes R-SAMPC to mitigate the impact of adverse factors and designs TVM to extract advantageous factors from the interference caused by R-SAMPC. Because the two modules in the serial structure will have contradictory effects (weakening and strengthening) to cancel out the effect. In order to fuse the functions of the two modules, SAM-TTT is designed as a structure of first parallel and then fusion. Unlike the serial structure, two modules do not interfere with each other in the parallel structure. Finally, in the fusion stage, the function of the two modules is combined. Dilated and grouped convolutions are employed to effectively capture multi-scale contextual information, enabling more precise fusion. Overall, SAM-TTT configures the parameters of SAM by broadening the Effect Distance between advantageous and adverse parameters in the model, partially addressing the semantic deficiency in SAM used for COD. Our contributions are summarized as follows:
\begin{itemize}
\item SAM-TTT is proposed for COD. SAM-TTT is a new SAM-based model that is the first to focus on expanding the Effect Distance between advantageous and adverse parameters, addressing the issue of semantic deficiency when applying SAM to COD.
\item Two efficient designs are proposed in SAM-TTT: R-SAMPC and TVM. R-SAMPC is a SAM parameter configuration module that mitigates the effects of adverse parameters in SAM in a train-free way. TVM extracts advantageous parameters to compensate for the parameter interference caused by R-SAMPC .
\item The performance of SAM-TTT is verified on COD benchmark datasets. SAM-TTT outperforms the existing SOTA methods. It can be concluded that SAM-TTT reaches the cutting-edge in this domain.
\end{itemize}
\begin{figure*}[h]
\centering{\includegraphics[width=0.9\linewidth]{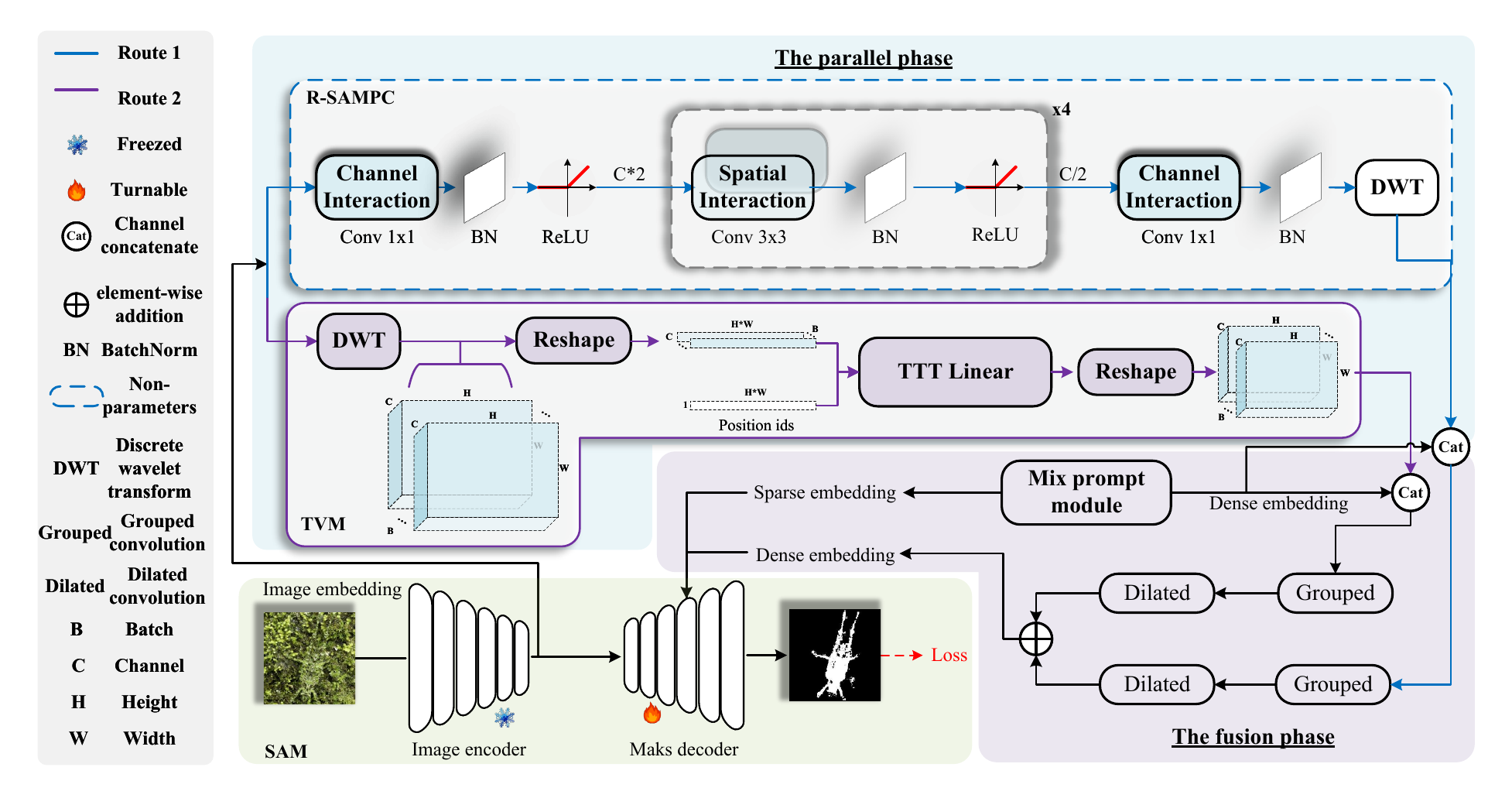}}
\caption{\small{\textbf{Overview of our SAM-TTT framework: the Reverse SAM Parameter Configuration Module (R-SAMPC) and the T-vision Module (TVM).} In the parallel phase, R-SAMPC and TVM operate independently, while in the fusion phase, the effectiveness of both modules is integrated.}}
\label{fig:SAM_TTT}
\end{figure*}
\section{Related Work}
\label{sec:related}
\subsection{Camouflaged Object Detection}
In nature, animals use camouflage as a defense strategy to evade predators~\cite{copeland1997models}.
COD~\cite{fan2023advances,xiao2024survey} is the process of identifying and segmenting camouflaged objects that blend into their surroundings. Many studies tackle this problem from multiple perspectives. They focus on gradient analysis~\cite{ji2023deep}, proactive template learning~\cite{asnani2023probed}, boundary detection~\cite{li2021uncertainty,sun2022boundary,zhu2021inferring,10834569}, depth cues~\cite{yu2024exploring,liu2024depth}, zoom-out strategies~\cite{pang2022zoom}, predation techniques~\cite{fan2021concealed,jia2022segment}, and other related approaches~\cite{yang2021uncertainty,zhang2021uncertainty}. Today's cutting-edge methods continue to explore these aspects more deeply. FSEL~\cite{sun2024frequency} highlights that previous methods overlook critical information found between high-frequency and low-frequency ranges. It integrates both global frequency and local spatial features to improve the detection of camouflaged objects. ZoomNeXt~\cite{pang2024zoomnext} mimics human behavior (i.e., zooming in and out) when observing vague images and videos. RISNet~\cite{wang2024depth} extends classic COD to agricultural domains. It employs multiscale receptive fields and integrates depth features to capture information about camouflage crops of varying sizes and spatial locations. It grounds the COD task in agricultural application. Due to the huge human effort required by the labels required for full supervision, some semi-supervised methods developed. Chen~\etal~\cite{chen2024just} proposed a novel point-based learning paradigm for the challenging COD task. GenSAM~\cite{hu2024relax} reasons visual prompts based on the semantic information from a generic text prompt of cross-modal large language mode.

\subsection{Camouflaged Object Detection based SAM}
As a novel foundation model, SAM is trained on a vast dataset and demonstrates impressive zero-shot capabilities. However, due to the domain gap between the data set trained by SAM and the data set of COD, the accuracy of SAM is inevitably reduced when it is applied to the field of COD~\cite{ji2023sam}. 
Similarly, Tang~\etal~\cite{tang2023can} directly studied COD and highlighted that SAM performs suboptimally on the COD task. One of the more specific embodiments is the lack of segmentation semantics ~\cite{chen2024sam}. This makes the internal segmentation of the camouflage target incomplete. He~\etal~\cite{he2024weakly} used sparse annotations from SAM as guidance for mask segmentation during model training.
SAM-Adapter~\cite{chen2023sam} integrates domain-specific information or visual prompts into the segmentation network using straightforward yet effective adapters. DSAM~\cite{yu2024exploring} refines the mask generated by SAM using depth cues to give semantic cues.
SAM-COD~\cite{chen2024sam} uses masks generated by an auxiliary network, which contain the initial semantics, to correct the masks predicted by SAM. COMPrompter~\cite{zhang2024COMPrompter} employs a mixed prompt approach, integrating both box prompts and boundary prompts to deliver more reliable semantic information. However, this method depends on the integration of an external auxiliary network or modules to tackle the challenge of internal semantic deficiency. While this approach can enhance performance, it also introduces additional parameters and computational complexity, which may impact efficiency. And it does not solve the problem of missing semantics of SAM applied to COD from the root, SAM itself.

\section{Methodology}
\subsection{Overall Architecture}
SAM-TTT consists of the Reverse SAM Parameter Configuration Module (R-SAMPC) and the T-visioner Module (TVM), shown in Figure~\ref{fig:SAM_TTT}. The image embedding is derived from the frozen image encoder of SAM. It possesses strong zero-shot capabilities, but it also contains knowledge that may not align with COD. To address this, SAM-TTT is proposed. It actively weakens knowledge that is adverse for COD and strengths knowledge that is advantageous. The network is structured with an initial parallel design followed by a fusion stage, aiming to maximize the distance between advantageous and adverse parameters. In the parallel phase, R-SAMPC and TVM act independently and do not interfere with each other, avoiding the possible effect cancellation of the weakening and the enhancing effects. In the fusion phase, the mix prompt method of COMPrompter is adopted. Features from R-SAMPC and TVM are fused into the hybrid prompts to guide the mask decoder. The issue of semantic deficiency of SAM applied to COD is alleviated.

\subsection{Reverse SAM Parameter Configuration Module}
The Reverse SAM Parameter Configuration Module (R-SAMPC) is designed to configure the parameters of SAM in a manner that increases the distance between adverse and advantageous features. In essence, R-SAMPC is a convolutional module that does not update parameters, functioning as a train-free method. It weakens the parameters in SAM that are not conducive to COD task, and obviously alleviates the lack of semantics of SAM applied to COD. Formally, this module can also be seen as a dropout, which only participates during training and not during inference. Traditional dropout either zeros out neurons or adds Gaussian noise. Unlike traditional dropout, R-SAMPC consists of a series of convolutional layers, Batch Normalization, and ReLU activation. R-SAMPC is only initialized during training and its parameters are not updated. So this module is trained in a train-free manner. Through R-SAMPC, image embedding disturbs the original parameters to a certain extent. R-SAMPC ($\mathcal{F}$) is expressed as follows:
\begin{align}
\mathcal{F} = CI_R 
\big(SI(CI_D\small(em_I\small))\big)
\end{align}
where $CI_D$ represents channel interaction using a 1x1 convolution with BatchNorm and ReLU, which doubles the number of channels. $SI$ denotes spatial interaction using the four layers of 3x3 convolution with BatchNorm and ReLU. $CI_R$ represents channel interaction using a 1x1 convolution with BatchNorm, which reduces the number of channels to one-half of the original. $em_I$ denotes embedding from image encoder. The function of $CI_D$ and $CI_R$ is to facilitate communication of the image embedding across the channel dimension. $SI$ enables simple communication over the spatial dimension. The simple convolutional combination in R-SAMPC is designed to introduce noise, with the aim of directly reducing the influence of adverse parameters in SAM. From this point of view, R-SAMPC can be seen as a mask for parameters. While previous works apply occlusion to the prediction mask~\cite{chen2024sam}, additional modules will be used to map the occlusion on the mask level to the parameter level. Other methods use the auxiliary network to correct mistakes (Figure~\ref{fig:dropout} (a), (b), (c)). Thus, R-SAMPC adds noise more directly and does not require additional subsequent computations (Figure~\ref{fig:dropout} (d)). From another point of view, adding random masks can force the model to focus on the overall objective by shifting its attention away from the highest parts~\cite{chen2024just}. Our R-SAMPC is exactly a parametric manifestation of this phenomenon. R-SAMPC is applied to $route 1$ in the parallel phase participating in SAM-TTT.
\subsection{T-visioner Module}
R-SAMPC weakens knowledge in image embedding that is detrimental to COD. At the same time, the knowledge that it benefits COD is also weakened inevitably. So, in another route in SAM-TTT, TVM is designed to emphasize the knowledge that SAM is beneficial for COD. Self-attention excels at handling long-range dependencies but comes with quadratic complexity. In contrast, traditional RNN layers have linear complexity, though their ability to model long-term context is constrained by the limited expressive capacity of their hidden states. Test-Time Training (TTT)~\cite{sun2024learning} layers solves this problem. The core concept is to treat the hidden state as a machine learning model. An instantiations of TTT is introduced, called TTT-linear, into $Route 2$ in the parallel phase to reinforce dominant features. The challenge that needs to be overcome is how to bring TTT, which is an RNN layer, to the field of computer vision. As shown in Figure ~\ref{fig:SAM_TTT}, the usage of DWT~\cite{he2023camouflaged} in COMPrompter is retained by us. DWT primarily captures diagonal high-frequency regions in an image, highlighting edges and subtle variations. DWT first extracts the high-frequency components of the image embedding and then adjusts the feature dimensions to comply with the requirements of the RNN layer. The prosess is as followed:
\begin{equation}
\text{reshape}: B \times C \times W \times H \longrightarrow B \times (W \times H) \times C
\end{equation}
Additionally, a positional encoding is generated . Its contents are integers from 0 to $W \times H$ - 1. The dimensions of it is as followed:
\begin{equation}
\text{Position encoding}: 1 \times(W \times H - 1) 
\end{equation}
After TTT-Liner, the shape will change the original dimension.
\begin{equation}
\text{reshape}: B \times (W \times H) \times C \longrightarrow  B \times C \times W \times H
\end{equation}
The role of TVM is to compensate for the interference caused by R-SAMPC on the advantageous parameters.

\section{Experiments}
\begin{table*}[h]
\centering
\renewcommand{\arraystretch}{1}
\setlength\tabcolsep{3.4pt}
\caption{\textbf{Quantitative results on three different datasets of CAMO, COD10K, and NC4K.} $\uparrow$ indicates the higher the score the better and $\downarrow$ indicates the lower the score the better. The highest and second-highest scores are \textbf{bolded}, while the third score is \underline{underlined}.}
\vspace{-6pt}
\begin{tabular}{c*{20}{c}}
\toprule
\multirow{2}{*}{Dataset} & \multirow{2}{*}{Metric} 
& \makecell[c]{SINet\\V2}     & \makecell[c]{Zoo\\mNet} 
& \makecell[c]{Seg\\MaR}        & \makecell[c]{DG\\Net} 
& \makecell[c]{MSC\\AFNet}  & \makecell[c]{Hit\\Net}   & \makecell[c]{PR\\Net} & \makecell[c]{RIS\\Net}  & \makecell[c]{Camo\\Focus} &   FSEL & PRBE & SAM  & \makecell[c]{SAMA\\dapter} 
& \makecell[c]{D\\SAM} & \makecell[c]{COMPr\\ompter}   
& \textbf{Ours}\\
\cmidrule(r){3-18} 
&& \makecell[c]{2022\\ \cite{fan2021concealed}}
& \makecell[c]{2022\\ \cite{pang2022zoom}}     
& \makecell[c]{2023\\ \cite{jia2022segment}}
& \makecell[c]{2023\\ \cite{ji2023deep}}   
& \makecell[c]{2023\\ \cite{liu2023mscaf}} 
& \makecell[c]{2023\\ \cite{hu2023high}} 
& \makecell[c]{2024 \\ ~\cite{hu2024efficient}} 
& \makecell[c]{2024\\ ~\cite{wang2024depth}} 
& \makecell[c]{2024 \\ \cite{khan2024camofocus}} 
& \makecell[c]{2024 \\ \cite{sun2024frequency}} 
& \makecell[c]{2024 \\ \cite{10814101}} 
& \makecell[c]{2023\\ \cite{kirillov2023segment}}     
& \makecell[c]{2023\\ \cite{chen2023sam}}
& \makecell[c]{2024 \\~\cite{yu2024exploring}} 
& \makecell[c]{2024 \\~\cite{zhang2024COMPrompter}}& -\\
\midrule
\multirow{4}{*}{\centering CAMO} 
& $F_\beta^\omega \uparrow$
& 0.743 & 0.752 & 0.742 & 0.769 
& 0.828 & 0.801 & 0.831 & 0.827  
& \textbf{0.842}  &\textbf{0.851} & 0.837 & 0.606 
& 0.765 & 0.794 & 0.819 & \underline{0.838}  \\
& $S_\alpha \uparrow$
& 0.820 & 0.820 & 0.815 & 0.839 
& 0.873 & 0.844 & 0.872 & 0.870  
& \underline{0.873} & \textbf{0.885} & \textbf{0.876} & 0.684 
& 0.847 & 0.832 & 0.853 & 0.868 \\
& $E_\phi \uparrow$
& 0.882 & 0.892 & 0.872 & 0.901 
& 0.929 & 0.902 & \underline{0.922} & \underline{0.922}  
& - &\textbf{0.942} & - & 0.687 
& 0.873 & 0.913 & 0.919 &\textbf{0.935} \\
& $M \downarrow$
& 0.070 & 0.066 & 0.071 & 0.057 
& 0.046 & 0.057 & 0.050 & 0.050  
& \textbf{0.043} &\textbf{0.040} & \underline{0.045} & 0.132 
& 0.070 & 0.061 & 0.054 &\underline{0.045} \\
\midrule
\multirow{4}{*}{\centering \makecell[c]{COD\\10K}} 
& $F_\beta^\omega \uparrow$
& 0.680 & 0.729 & 0.724 & 0.693 
& 0.775 & 0.798 & \textbf{0.803} & \underline{0.802} 
& \underline{0.802} &  \textbf{0.805} & 0.793 & 0.701 
& 0.801 & 0.760 & 0.779 & \textbf{0.805} \\
& $S_\alpha \uparrow$
& 0.815 & 0.838 & 0.833 & 0.822 
& 0.865 & 0.868 & 0.873 & 0.873
& 0.873 &\textbf{0.877} & 0.867 & 0.783 
& \textbf{0.883} & 0.846 & 0.861 &\underline{0.874} \\
& $E_\phi \uparrow$
& 0.887 & 0.911 & 0.895 & 0.896 
& 0.927 & 0.932 &\underline{0.935} & 0.930 
& - & \textbf{0.938} & - & 0.798 
& 0.918 & 0.921 & 0.933 & \textbf{0.942} \\
& $M \downarrow$
& 0.037 & 0.029 & 0.033 & 0.033 
& \underline{0.024} & \underline{0.024} & 0.025 & 0.028 
& \textbf{0.021} & \textbf{0.023} & \textbf{0.021} & 0.049 
& 0.025 & 0.033 & 0.026 & 0.027 \\
\midrule
\multirow{4}{*}{\centering NC4K}
& $F_\beta^\omega \uparrow$
& 0.770 & 0.784 & 0.781 & 0.784 
& 0.839 & 0.825 & 0.832 & 0.823 
&\textbf{0.853} & 0.836 & \textbf{0.845} & 0.696 
& -  & 0.826 & 0.840 & \underline{0.837} \\
& $S_\alpha \uparrow$
& 0.847 & 0.853 & 0.841 & 0.857 
& \textbf{0.887} & 0.870 & \underline{0.885} & 0.879 
& \textbf{0.889} & \textbf{0.887} & \textbf{0.887} & 0.767
& -  & 0.871 & 0.880 & 0.884 \\
& $E_\phi \uparrow$
& 0.903 & 0.912 & 0.905 & 0.911 
& \underline{0.935} & 0.921 & \underline{0.935} & 0.927 
& - & \textbf{0.940} & - & 0.776 
& - & 0.932 & \underline{0.935} &\textbf{0.943} \\
& $M \downarrow$
& 0.048 & 0.043 & 0.046 & 0.042 
& 0.032 & 0.039 & \textbf{0.029} & 0.033 
& \textbf{0.030} & \textbf{0.028} & 0.031 & 0.078 
& - & 0.040 & 0.036 & 0.031 \\
\bottomrule
\end{tabular}
\label{tab:tab_COD}
\end{table*}
\begin{figure*}[h]
\centering{\includegraphics[width=0.85\linewidth]{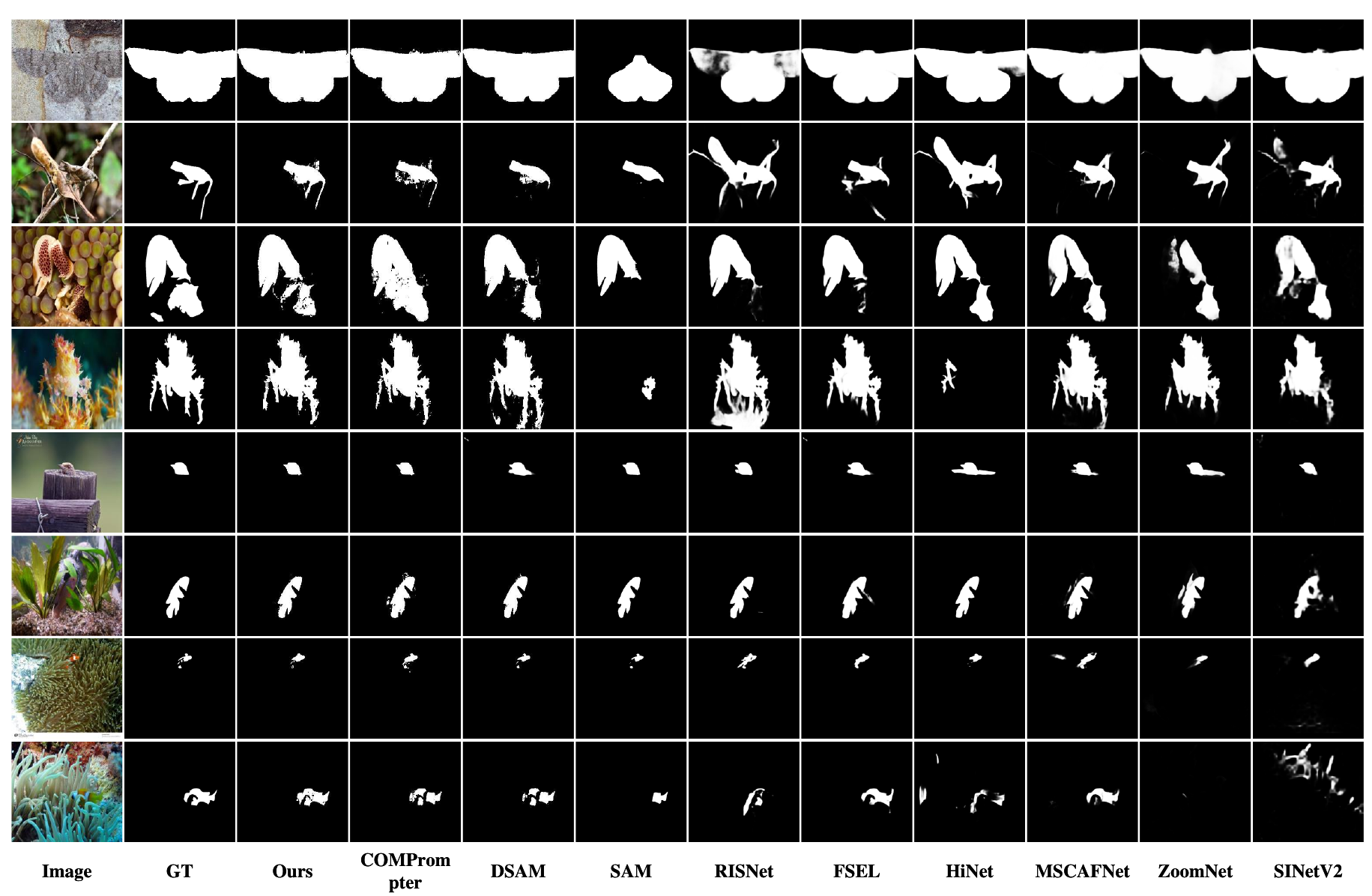}}
\vspace{-10pt}
\caption{\small{\textbf{Comparison of our SAM-TTT and other methods on different types of samples.} (Better to zoom in.)}}
\label{fig:comparison}
\end{figure*}
\subsection{Datatset}
Experiments are conducted on three widely used datasets, namely CAMO~\cite{le2019anabranch}, COD10K~\cite{fan2021concealed}, and NC4K~\cite{lv2021simultaneously}. This aims to evaluate the impact of SAM-TTT on the task of COD. CAMO contains a total of 1250 images, which are randomly divided into a training set of 1000 images and a test set of 250 images. COD10K includes 5066 images, with 3040 used for training and 2026 for testing. NC4K is a relatively large dataset, consisting of 4121 images in total. It is utilized as a test set in experiments to assess the generalization capability of SAM-TTT.
Following Fan \etal~\cite{fan2021concealed}, our study adopts that the dataset is composed of the train datasets of COD10K and CAMO, which are 3040 images and 1000 images, respectively. The remaining images of COD10K and CAMO and the entire NC4K dataset are used as the test dataset.
\subsection{Experimental Setup}
\textbf{Implementation Details.}
SAM-TTT is implemented using PyTorch, employing the Adam optimizer with a learning rate of $1e^{-5}$. The model is trained for 290 epochs to achieve optimal performance, taking around 24 hours to complete on an NVIDIA 3080TI GPU with a batch size of 16. All input images are resized to $1024 \times 1024$ using bilinear interpolation, scaling them up or down as needed. In addition, the input image data are truncated and normalized. This ensures the pixel values are in the appropriate range while maintaining the relative distribution relationship of the data.

\textbf{Evaluation Metrics.}
Four metrics from COD10K~\cite{fan2021concealed} are adopted. These four metrics are commonly used and well-established in the field of COD: structure measure ($S_\alpha$)~\cite{fan2017structure}, weighted F-measure ($F_\beta^\omega$)~\cite{margolin2014evaluate}, mean enhanced-alignment measure ($E_\phi$)~\cite{fan2018enhanced}, and mean absolute error ($M$). The structure measure quantifies the structural similarity between the predicted results and the actual segmented regions. The weighted F-measure combines precision and recall, and weights them. The enhanced-alignment measure evaluates the prediction result by comparing the alignment relationship between the predicted value and the actual value. The mean absolute error is a measure of the mean absolute error between the predicted value and the true value.
\begin{table}[t]
\centering
\renewcommand{\arraystretch}{1}
\setlength\tabcolsep{9pt}
\caption{Comparison of Network Complexity. The parameters in parentheses represent the training parameters.}
\label{tab:complexity}
\begin{tabular}{lccc}
\toprule
Models & Input size & Param (M) & Speed (fps) \\
\midrule
SINetV2   & 352×352     & 26.98 (26.98) & 50 \\
DGNet     & 352×352     & 21.02 (21.02) & 40 \\
FSEL      & 416×416     & 67.13 (67.13) & 31 \\
SAM       & 1024×1024   & 91 (91)       & 2  \\
SAM-TTT   & 1024×1024   & 96.32 (6.65)  & 3  \\
\bottomrule
\end{tabular}
\end{table}
\subsection{Comparisons with Cutting-Edge Methods}
Here, data from top-performing models over the past three years is selected. To ensure a comprehensive comparison, both SAM-based models and non-SAM methods are included. SAM-TTT is compared with SAM~\cite{kirillov2023segment} and other existing COD algorithms, such as SINetV2~\cite{fan2021concealed}, 
ZoomNet~\cite{pang2022zoom}, SegMaR~\cite{jia2022segment},
DGNet~\cite{ji2023deep}, MSCAF-Net~\cite{liu2023mscaf}, HitNet~\cite{hu2023high}, PRNet~\cite{hu2024efficient}, RISNet~\cite{wang2024depth}, PNet~\cite{zhang2024learning}, CamoFocus~\cite{khan2024camofocus}, FSEL~\cite{sun2024frequency},
SAM-Adapter~\cite{chen2023sam},
DSAM~\cite{yu2024exploring},
COMPrompter~\cite{zhang2024COMPrompter}.
The predictions of the competitors are disclosed by the authors or generated by models retrained using open-source code. The comparison results are presented in Table~\ref{tab:tab_COD}.

\begin{table}[h]
\centering
\renewcommand{\arraystretch}{1}
\setlength\tabcolsep{7.2pt}
\caption{\small{\textbf{Ablation study results for each module of the proposed SAM-TTT on COD datasets.} $P$ represents the average value of positive metrics, while $N$ represents the average value of negative metrics. $\uparrow$ indicates the higher the score the better and $\downarrow$ indicates the lower the score the better.}}
\begin{tabular}{cccccc}
\toprule
Dataset & Metric & M1& M2& M3* & M3\\
\midrule
\multirow{4}{*}{\centering CAMO} 
& $F_\beta \uparrow$
& 0.819 &  0.836 & 0.837 & 0.838 \\
& $S_\alpha \uparrow$
& 0.853 & 0.864 & 0.866 & 0.868\\
& $E_\phi \uparrow$
& 0.919 & 0.929 & 0.933 & 0.935 \\
& $M \downarrow$
&  0.054 & 0.047 & 0.047 & 0.045 \\
\midrule
\multirow{4}{*}{\centering COD10K} 
& $F_\beta \uparrow$
&  0.779 &  0.799 &  0.808 &  0.805 \\
& $S_\alpha \uparrow$
& 0.861 &  0.869 &  0.873 &  0.874 \\
& $E_\phi \uparrow$
&  0.933 &  0.937 &  0.941 &  0.942 \\
& $M \downarrow$
&  0.026 &  0.027 &  0.026 & 0.027 \\
\midrule
\multirow{4}{*}{\centering NC4K}
& $F_\beta \uparrow$
&  0.840 &  0.833 & 0.839 & 0.837 \\
& $S_\alpha \uparrow$
&  0.880 &  0.881 &  0.885 &  0.884 \\
& $E_\phi \uparrow$
&  0.935 &  0.939 &  0.942 &  0.943 \\
& $M \downarrow$
&  0.036 & 0.031 &  0.030 &  0.031 \\
\midrule
\multirow{2}{*}{\centering Average}
& $P \uparrow$ & 0.869 &  0.876 &  0.880 &  0.881\\
& $N \downarrow$ &  0.0387 &  0.0350 &  0.0343 &  0.0343\\
\bottomrule
\end{tabular}
\label{tab:tab_abl}
\end{table}
\textbf{Efficiency analysis.}
Table~\ref{tab:complexity} compares parameter counts and inference speed. Although SAM-TTT has a large total parameter size (96.32M), its trainable parameters are only 6.65M—about one-tenth of FSEL’s. This keeps computational overhead low while outperforming fully supervised methods under similar complexity.

\textbf{Quantitative Results.}
The quantitative comparison results of the proposed algorithm and the other 15 methods on the three datasets are shown in Table~\ref{tab:tab_COD}. The 15 methods include both SAM-based models and non-SAM-based models. No one method has a completely significant advantage. Given the similar and competitive accuracy, the top three scores are highlighted across 12 metrics on three datasets. There exists some same score in top three score. The identical top scores reflect the intense level of competition. Among the top 2 scores, half of the differences are less than 0.3\%. The table highlights that among the top three scores across 12 metrics, 6 metrics have overlapping data. SAM-TTT secures top-three positions in 8 out of 12 metrics, indicating a certain level of advancement. On CAMO, SAM-TTT performs slightly worse), likely due to enhanced generalization from SAM knowledge and R-SAMPC perturbations at the expense of learning ability. The high train-test ratio (1000:250) may further amplify this effect.

\begin{figure}[h]
\centering{\includegraphics[width=0.8\linewidth]{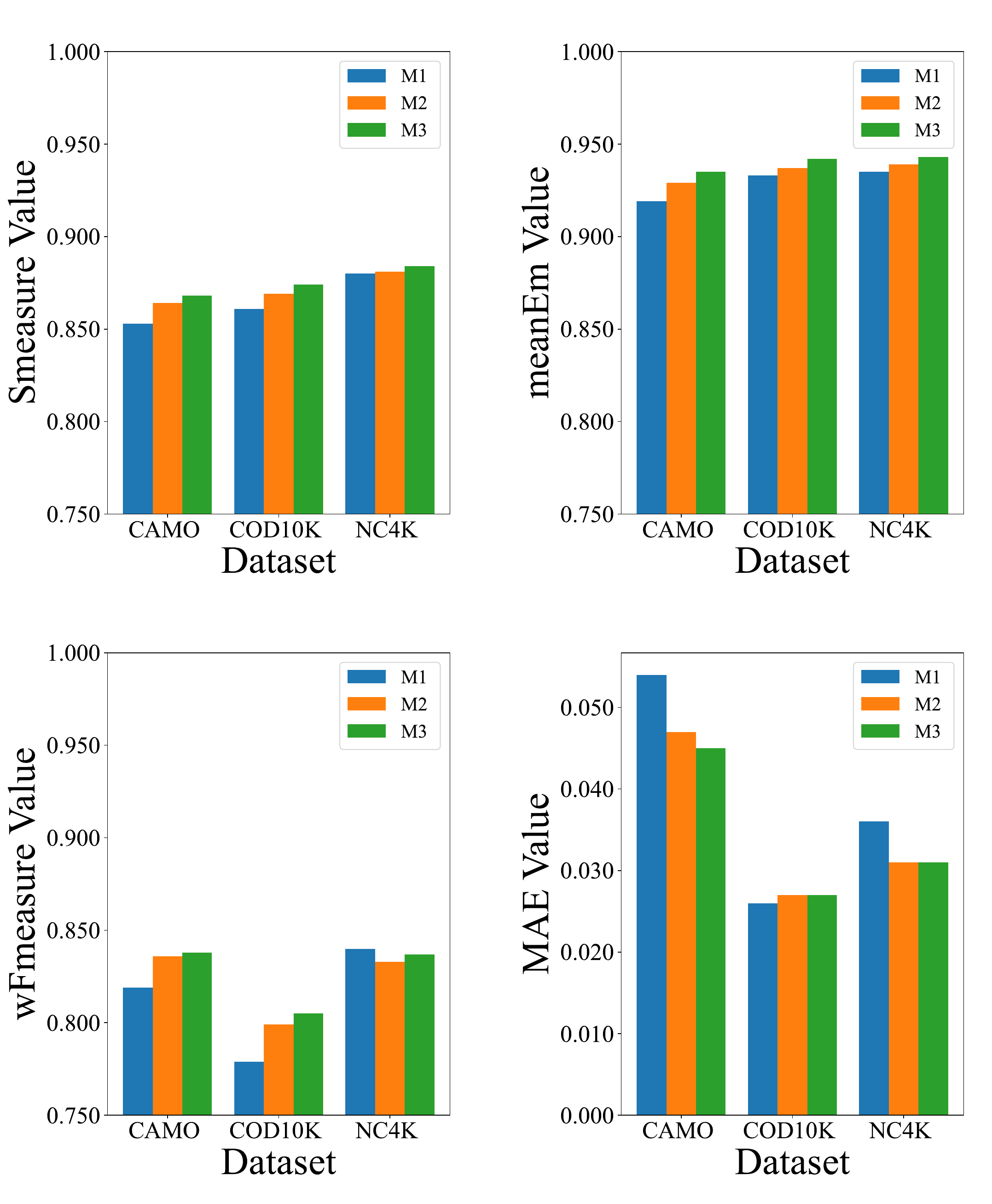}}
\caption{\small{\textbf{Visualization of ablation experiments.} $M1$ denotes baseline, COMPrompter. $M2$ denotes baseline + R-SAMPC. $M3$ denotes baseline + R-SAMPC + TVM, which is SAM-TTT.}}
\label{fig:abla}
\end{figure}

\begin{figure*}[h]
\centering{\includegraphics[width=0.8\linewidth]{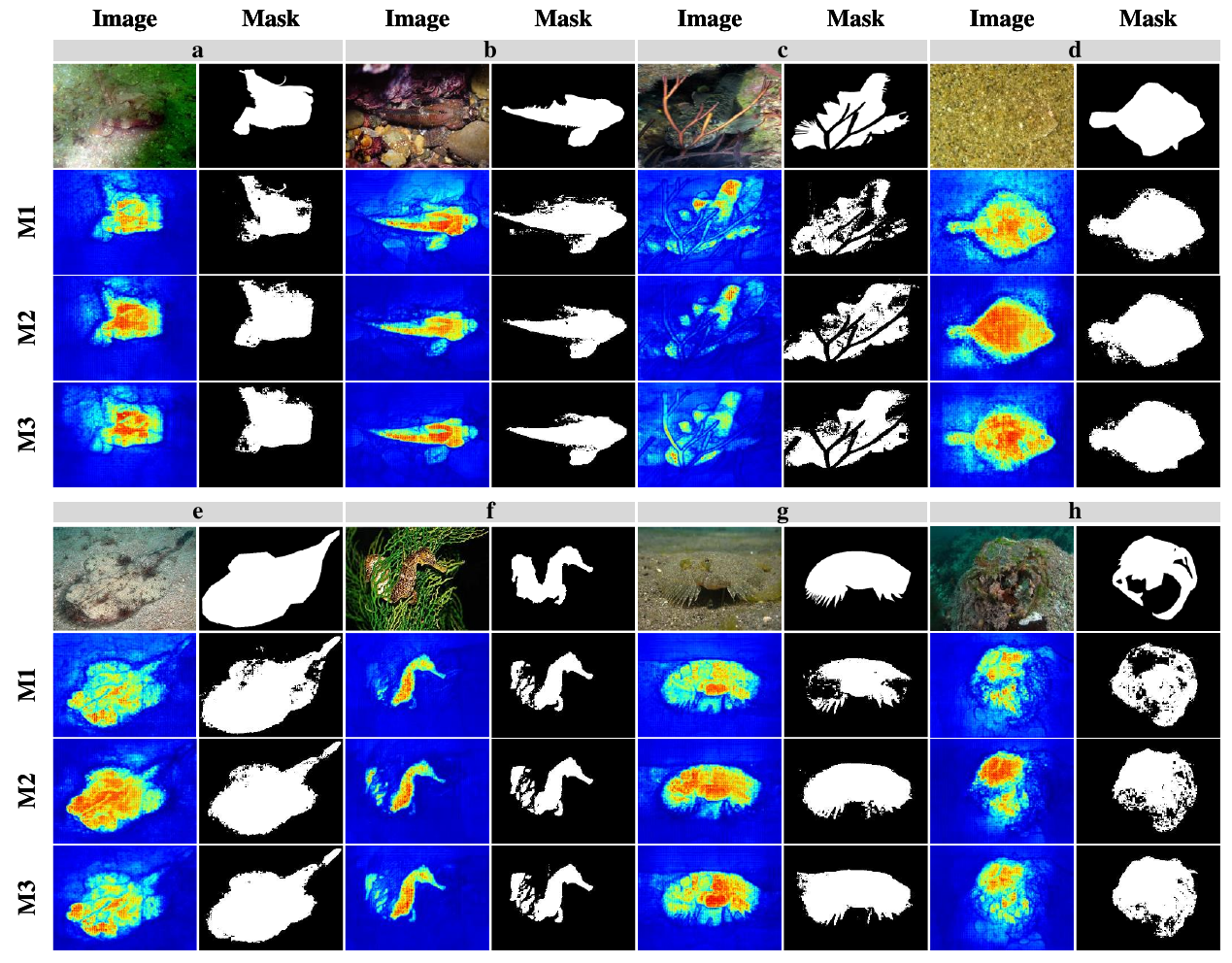}}
\caption{\small{\textbf{Group-wise feature map, mask, image, and ground truth comparison based on ablation settings.} $M1$ denotes baseline, COMPrompter. $M2$ denotes baseline + R-SAMPC. $M3$ denotes baseline + R-SAMPC + TVM, which is SAM-TTT.}}
\label{fig:feature}
\end{figure*}

\textbf{Qualitative Results.}
The visual comparisons of different methods on COD datasets are presented in Figure~\ref{fig:comparison}. The masks from various perspectives are selected, including tiny objects (Rows 5 and 7), small objects (Rows 6 and 8), large objects (Rows 1 and 3), occluded objects (Rows 5 and 6), and objects with fine structures (Row 2 and 4). In Rows 2 and 4, the details of the objects are clearly shown through SAM-TTT. In contrast, the other methods exhibit either over-segmentation or under-segmentation, leading to blurred details. The illustrated results show that SAM-TTT demonstrates a more detailed handling of fine features. Under occlusion, the model requires more semantic knowledge. In Rows 6, 7, and 8, SAM-TTT effectively distinguishes occlusions from objects and demonstrates more accurate segmentation. SAM-TTT provides a more comprehensive semantic understanding when addressing occlusion. A similar background can be beneficial for camouflaged objects. In Row 4, the background is similar to the object, making it challenging for many methods to detect the boundaries between the target and the background. However, SAM-TTT demonstrates superior discrimination in such cases, as shown in Row 4.
\subsection{Ablation study}
\begin{table*}[h]
\centering
\renewcommand{\arraystretch}{0.9}
\setlength\tabcolsep{2pt}
\caption{\small{\textbf{Ablation study results for R-SAMPC.} $L0$ denotes COMPrompter. $L1$ denotes baseline + a layer of convolutional combinations. $L2$ denotes two layers. $L3$ denotes three layers. $L4$ denotes four layers. $L4+\epsilon$ denotes $L4$ + channel scaling. $P$ represents the average value of positive metrics, while $N$ represents the average value of negative metrics. The best data is marked \textbf{bold}, and the second best data is \underline{underlined}. $\uparrow$ indicates the higher the score the better and $\downarrow$ indicates the lower the score the better.}}
\begin{tabular}{ccccccccccccccccccc|cc}
\toprule
\multirow{2}{*}{Setting} & \multicolumn{6}{c}{CAMO} & \multicolumn{6}{c}{COD10K} & \multicolumn{6}{c}{NC4K} & \multicolumn{2}{c}{Average} \\
\cmidrule(r){2-7} \cmidrule(r){8-13} \cmidrule(r){14-19} \cmidrule(r){20-21} 
& 
$S_\alpha \uparrow$  & $F_\beta^\omega \uparrow$ & $E_\phi \uparrow$ & $F_m \uparrow$ & $E_x \uparrow$ & $M \downarrow$&
$S_\alpha \uparrow$  & $F_\beta^\omega \uparrow$ & $E_\phi \uparrow$ & $F_m \uparrow$ & $E_x \uparrow$ & $M \downarrow$ &
$S_\alpha \uparrow$  & $F_\beta^\omega \uparrow$ & $E_\phi \uparrow$ & $F_m \uparrow$ & $E_x \uparrow$ & $M \downarrow$ & $P$ & $N$\\
\midrule
$L0$
& 0.853 & 0.819 & 0.919 & 0.843 & 0.931 & 0.054 
& 0.861 & 0.779 & 0.933 & 0.806 & 0.945 & 0.026  
& 0.880 & 0.840 & 0.935 & 0.862 & 0.946 & 0.036 & 0.8768 & 0.0387\\
$L1$
& 0.863 & 0.834  & 0.928 & 0.855 & 0.938 & 0.048 
& 0.868 & 0.797  & 0.937 & 0.823 & 0.948 & 0.027 
& 0.880 & 0.831  & 0.938 & 0.853 & 0.948 & 0.031 & 0.8827 & 0.0353\\
$L2$
& 0.861 & 0.831  & 0.927 & 0.852 & 0.937 & 0.049 
& 0.868 & 0.797  & 0.937 & 0.821 & 0.948 & 0.027 
& 0.881 & 0.831  & 0.939 & 0.852 & 0.948 & 0.032 & 0.8820 & 0.0360\\
$L3$
& 0.864 & 0.834  & 0.929 & 0.855 & 0.940 & 0.048 
& 0.869 & 0.799  & 0.937 & 0.824 & 0.948 & 0.027 
& 0.881 & 0.833  & 0.939 & 0.854 & 0.949 & 0.031 & \underline{0.8837} & 0.0353\\
$L4$
& 0.864 & 0.835 & 0.930 & 0.855 & 0.94 & 0.047  
& 0.869 & 0.799 & 0.938 & 0.823 & 0.948 & 0.027  
& 0.881 & 0.832 & 0.939 & 0.854 & 0.949 & 0.031 & \underline{0.8837} & \textbf{0.0350}\\
$L5$
& 0.864 & 0.834 & 0.929 & 0.854 & 0.939 & 0.048  
& 0.868 & 0.797 & 0.936 & 0.821 & 0.946 & 0.027  
& 0.881 & 0.832 & 0.938 & 0.852 & 0.948 & 0.031 & 0.8826 & 0.0353\\
$L4+\epsilon$
& 0.864 & 0.836  & 0.929 & 0.857 & 0.94 & 0.047 
& 0.869 & 0.799  & 0.937 & 0.824 & 0.947 & 0.027 
& 0.881 & 0.833  & 0.939 & 0.855 & 0.949 & 0.031 & \textbf{0.8839} & \textbf{0.0350}\\
\bottomrule
\end{tabular}
\label{tab:tab_abl_R-SAMPC}
\end{table*}
\begin{table}[h]
\centering
\caption{\small{\textbf{Degradation of R-SAMPC for Adverse Parameters ($\times 10^{-3}$)}. Roman numerals (e.g., I, II) indicate different adverse parameters. Larger absolute values are better.}}
\label{tab:rsampc_degradation}
\renewcommand{\arraystretch}{0.9}
\setlength\tabcolsep{1pt}
\begin{tabular}{lccccc}
\toprule
\textbf{Model-SAM} & \textbf{I} & \textbf{II} & \textbf{III} & \textbf{IV} & \textbf{V} \\
\midrule
\textbf{COMPrompter(M)} & -3.2 & -0.9 & -0.5 & -0.3 & -0.3 \\
\textbf{M+R-SAMPC} & -3.9 & -2.2 & -1.8 & -1.1 & -1.1 \\
\textbf{R-SAMPC Gain} & -0.7 & -1.3 & -1.3 & -0.8 & -0.8 \\
\textbf{Relative Gain} & -21.87\% & -144.44\% & -260.00\% & -266.67\% & -266.67\% \\
\bottomrule
\end{tabular}
\end{table}
Ablation experiments are conducted to evaluate the effectiveness of R-SAMPC and TVM. Specifically, R-SAMPC serves as a prerequisite for TVM. Four models are designed to demonstrate the effectiveness of R-SAMPC and TVM. COMPrompter is chosen as the baseline model, referred to as $M1$. $M2$ adds R-SAMPC and the corresponding fusion phase to $M1$, while $M3$ incorporates TVM and the corresponding fusion phase based on $M2$. In addition, $M3*$ replaces TTT with Mamba~\cite{gu2023mamba} to assess the effectiveness of TTT. The quantitative results of different models in SAM-TTT are provided, shown in Table~\ref{tab:tab_abl}. In order to better observe the accuracy gap between models, Table~\ref{tab:tab_abl} is visualized as Figure~\ref{fig:abla}. To observe the effect of R-SAMPC and TVM, the feature visualizations of $M1$, $M2$, $M3$ are compared with the final maks of this model, displayed in Figure~\ref{fig:feature}. Finally, to confirm the rationality of R-SAMPC, an ablation experiment is performed on the rationality of R-SAMPC structure, which is shown in Table~\ref{tab:tab_abl_R-SAMPC}.

\textbf{Effectiveness of R-SAMPC.}
The effectiveness of R-SAMPC is demonstrated by the gap between $M1$ and $M2$. As shown in Figure~\ref{fig:abla}, $M2$ achieves an overall improvement compared to $M1$, with particularly notable gains in the three positive metrics. The effectiveness of R-SAMPC is analyzed based on average metrics. $M2$ achieves an average improvement of 0.6\% in $S_\alpha$, 1.0\% in $F_\beta^\omega$, 0.6\% in $E_\phi$ across three datasets. The negative metric, $M$ is reduced by 0.37\%. Overall, $M2$ shows a 0.7\% improvement in positive metrics compared to $M1$. To clarify the role of R-SAMPC, it is analyzed from the perspective of feature maps. Figure~\ref{fig:feature} lists a total of 8 groups of feature maps and mask comparison maps. From the feature maps of $M1$ in groups a, e, g and h, it is obvious that there are low response areas caused by semantic errors inside the object in $M1$. Part of this low response is reflected in the final prediction map being abnormal hollowing inside the object (see $M1$'s masks of groups a, e, g and h). $M2$ uses R-SAMPC, and this phenomenon is alleviated in the feature map. The hollowing in the corresponding final prediction map is also partially compensated (see $M2$ line of groups a, e, g and h). Some hollow segments are close to the edge, so R-SAMPC can also help to segment the edge (see group b, c, d and f). In addition, the structural soundness of the R-SAMPC module is demonstrated in Table~\ref{tab:tab_abl_R-SAMPC}. The module design gap of R-SAMPC is more subtle, and six more comprehensive indicators are adopted. The metrics added are mean F-measure ($F_m$) and max enhanced-alignment measure ($E_x$). The pioneering use of a single convolutional layer as dropout gives the largest improvement, which is 0.59\% in terms of average positive index and 0.33\% in terms of MAE. When the number of convolutional layers reaches four ($L4$), the index reaches its peak. By adding the channel scaling, the average positive accuracy increases by 0.02\% compared with $L4$. Therefore, the rationality and effectiveness of the R-SAMPC are demonstrated, confirming its suitability. 
To validate the mitigation effect of R-SAMPC on adverse parameters, we selected five parameters where COMPrompter shows reduced degradation compared to SAM, and conducted a quantitative comparison with the mitigation achieved by R-SAMPC. Details are provided in Table~\ref{tab:rsampc_degradation}.

\begin{table}[h]
\centering
\renewcommand{\arraystretch}{0.9}
\setlength\tabcolsep{1pt}
\caption{\small{\textbf{Enhancement of TVM for Advantageous Parameters ($\times 10^{-3}$)}. Roman numerals indicate different parameters. Larger absolute values are better.}}
\begin{tabular}{lccccc}
\toprule
\textbf{Models-SAM} & \textbf{I} & \textbf{II} & \textbf{III} & \textbf{IV} & \textbf{V} \\
\midrule
\textbf{COMPrompter (M)} & 0.171 & 0.103 & 0.027 & 0.028 & 0.045 \\
\textbf{M+R-SAMPC}       & 0.366 & 0.048 & 0.004 & 0.032 & 0.066 \\
\textbf{M+R-SAMPC+TVM}   & 0.465 & 0.073 & 0.027 & 0.054 & 0.070 \\
\textbf{TVM Gain}        & +0.099 & +0.025 & +0.023 & +0.022 & +0.004 \\
\textbf{Relative Gain}   & +27.05\% & +52.08\% & +575.00\% & +68.75\% & +6.06\% \\
\bottomrule
\label{tab:tvm_gain}
\end{tabular}
\end{table}
\textbf{Effectiveness of TVM.}
The effectiveness of TVM is demonstrated by the gap between $M3$ and $M2$. It can be obtained from Figure~\ref{fig:abla} that for the positive indicators, the improvement brought by $M3$ is not as large as the improvement brought by $M2$ to $M1$. However, the improvement of $M3$ also comprehensively exceeds all the accuracy of $M2$. The effectiveness of TVM is analyzed based on average metrics. $M3$ achieves an average improvement of 0.4\% in $S_\alpha$, 0.4\% in $F_\beta^\omega$, 0.5\% in $E_\phi$ across three datasets. Specifically, the $E_\phi$ has been improved by 0.6\%, 0.5\%, and 0.4\% in the three datasets. The comparison between $M3$ and $M3*$ demonstrates that TTT achieves a higher positive metrics compared to Mamba. This indicates that TTT has a stronger ability to focus on beneficial features. The role of TVM is to compensate for the side effects introduced by R-SAMPC. This side effect weakens the advantageous parameters. It can be observed from the feature map of $M1$ and $M2$ that R-SAMPC makes the overall response more balanced, though it introduces some erroneous responses. TVM extracts features on the basis of R-SAMPC to correct the error response (see $M3$ line of d in Figure~\ref{fig:feature}). To validate the enhancement effect of TVM on advantageous parameters, we selected five parameters where COMPrompter outperforms SAM, and performed a quantitative comparison of the improvements brought by TVM. Details are provided in Table~\ref{tab:tvm_gain}.

\section{Conclusion}
\label{sec:Conclusion}
The paper proposes SAM-TTT, a novel SAM-based network designed to address the semantic deficiency that occurs when applying SAM to COD. SAM-TTT consists of R-SAMPC and TVM. It offers a new approach for applying SAM to COD, widening the Effect Distance between advantageous and adverse parameters. The state-of-the-art performance of SAM-TTT is demonstrated over 15 cutting-edge methods across multiple datasets. SAM-TTT is a groundbreaking step towards non-parametric weakening of adverse parameters in SAM, and serves as a foundational step in introducing Test-Time Training to computer vision. Improving the combination of weakening adverse ones and emphasizing advantageous ones remains a key area for further exploration. Broadly, this concept could extend to other large model applications in downstream tasks.
\begin{acks}
This work was supported in part by the National Key Research and Development Program Project of China [grant no. 2024YFC3306902], in part by the National Natural Science Foundation of China [grant nos. U24A20242, 62475241, 62472312], and in part by the Zhejiang Provincial Natural Science Foundation [grant no. LDT23F02024F02].
\end{acks}

\bibliographystyle{ACM-Reference-Format}
\balance
\bibliography{sample-base}

\end{document}